\documentclass{article} % For LaTeX2e
\usepackage[preprint]{colm2025_conference}

\usepackage{microtype}
\usepackage{hyperref}
\usepackage{url}
\usepackage{booktabs}

\usepackage{lineno}
\usepackage{amsmath}
\usepackage{amssymb}
\usepackage{mathtools}
\usepackage{amsthm}
\usepackage{graphicx}
\usepackage{subfigure}
\usepackage{multirow}
\usepackage{graphicx}
\usepackage{wrapfig}
\usepackage{multirow}
\usepackage{booktabs}
\usepackage{authblk}  % For author and affiliation formatting
\usepackage{amsmath}  % Optional, for better typesetting
\usepackage{authblk}

\usepackage[capitalize,noabbrev]{cleveref}

\theoremstyle{plain}

\theoremstyle{definition}

\theoremstyle{remark}

\definecolor{darkblue}{rgb}{0, 0, 0.5}
\hypersetup{colorlinks=true, citecolor=darkblue, linkcolor=darkblue, urlcolor=darkblue}

\title{HyperRAG: Enhancing Quality-Efficiency Tradeoffs in Retrieval-Augmented Generation with Reranker KV-Cache Reuse}

\author{
\textbf{Yuwei An}$^{1}$, 
\textbf{Yihua Cheng}$^{2}$, 
\textbf{Seo Jin Park}$^{3}$, 
\textbf{Junchen Jiang}$^{2}$ \\
$^{1}$Carnegie Mellon University \\
$^{2}$University of Chicago \\
$^{3}$University of Southern California
}
\begin{document}

\ifcolmsubmission
\linenumbers
\fi

\maketitle

\begin{abstract}
Retrieval-Augmented Generation (RAG) has emerged as a powerful paradigm for enhancing the performance of large language models (LLMs) by integrating external knowledge into the generation process. A key component of RAG pipelines is the reranker, which selects the most relevant documents from a pool of retrieved candidates and significantly improves the quality of the generated responses. While rerankers refine the selection of retrieved documents in RAG pipelines, they introduce computational challenges that hinder high throughput and low latency. To address this problem, we propose HyperRAG, a system that optimizes the trade-off between quality and efficiency in RAG pipelines by leveraging KV-cache reuse for efficient reranker inference. By reusing document-side KV-cache, HyperRAG achieves both high-quality generation and system-level efficiency. To fully realize the benefits of KV-cache reuse, HyperRAG incorporates a range of system-level optimizations designed to enhance efficiency and scalability. Experiments show that HyperRAG achieves a 2–3× throughput improvement with decoder-only rerankers while also delivering higher downstream performance compared with traditional RAG service.
\end{abstract}

\section{Introduction}

Retrieval-Augmented Generation (RAG) \citep{gao2024retrievalaugmentedgenerationlargelanguage} has emerged as a powerful paradigm that enhances the performance of large language models (LLMs) by incorporating external knowledge into the generation process. RAG systems typically follow a two-stage pipeline: retrieval and generation. Among these, the retrieval stage plays a critical role as the relevance and quality of the retrieved documents significantly influence the final output. Selecting the most appropriate documents from a large corpus determines how effectively the model can respond to queries, especially in open-domain settings.

Compared with the coarse-grained retrieval which is based on dense embedding similarity over the vector index \citep{izacard2022unsuperviseddenseinformationretrieval, chen2024bgem3embeddingmultilingualmultifunctionality}, the reranker performs fine-grained selection by scoring candidates with richer contextual understanding, leading to more relevant and concise results \citep{ reimers-2019-sentence-bert, Gao_2022}. However, the reranking step introduces additional computational overhead and latency which can become a bottleneck in high-throughput RAG systems. This issue is further exacerbated by the adoption of large language model (LLM)-based rerankers \citep{chen2024bgem3embeddingmultilingualmultifunctionality, ma2023finetuningllamamultistagetext, pradeep2023rankzephyr}. On one hand, these rerankers are fine-tuned from powerful pre-trained generative models and achieve state-of-the-art performance on various retrieval tasks, such as passage ranking and document selection. On the other hand, they are computationally intensive and introduce substantial latency, especially when handling large batches of queries. As a result, despite their strong performance, LLM-based rerankers are often impractical for real-time RAG applications.

So the core challenge we aim to address is:

\vspace{0.1em}
\textit{How can we optimize the trade-off between generation quality and efficiency in RAG systems, especially with large-scale LLM-based rerankers, to deliver effective results without significantly compromising system throughput?}

To address this challenge, we propose HyperRAG, a system design that leverages efficient KV-cache management to optimize the quality-efficiency trade-off in the RAG system. Specifically, HyperRAG introduces a KV-cache storage mechanism that stores the KV-cache of all document chunks. When performing query and document reranking, HyperRAG efficiently loads the precomputed KV-cache, eliminating the need for recomputation and reducing latency while maintaining high generation quality. This mechanism shifts the bottleneck from GPU computation to SSD storage and transfer bandwidth, effectively balancing the workload and enabling more efficient utilization of system resources.

The main contribution of HyperRAG includes:
\begin{itemize}
     \item \textbf{Highlighting the potential of decoder based reranking}:  HyperRAG emphasizes the critical importance of rerankers in improving generation quality within RAG pipelines. Our empirical findings show that incorporating powerful decoder-based rerankers significantly enhances downstream performance, revealing the untapped potential of reranking in practical RAG deployments.
     
    \item \textbf{Enabling efficient decoder-based reranking via KV-cache reuse}: To address the computational inefficiency problem associated with decoder-based rerankers, HyperRAG introduces a  KV-cache reuse mechanism. By caching the document-side key/value pairs, the reranker only needs to process the query portion during inference. This shifts the bottleneck from GPU compute to underutilized resources such as NVM storage and PCIe bandwidth, enabling high-throughput inference while maintaining strong generation quality.

\end{itemize}

Our experiments demonstrate that HyperRAG maintains high-quality generation while delivering a 2–3× throughput improvement during RAG service. This design paves the way for a new pattern for deploying RAG systems.
\section{Background}
In this section, we introduce the background and related work underlying the design of HyperRAG.

% , focusing primarily on Retrieval-Augmented Generation (RaG) and KV-cache reuse strategies for efficient LLM deployment.

\subsection{Retrieval-Augmented Generation}
Retrieval-Augmented Generation (RAG) \citep{ram2023incontextretrievalaugmentedlanguagemodels, lewis2021retrievalaugmentedgenerationknowledgeintensivenlp,  asai2023selfraglearningretrievegenerate, khandelwal2020generalizationmemorizationnearestneighbor,
jin2024longcontextllmsmeetrag, shao2024scalingretrievalbasedlanguagemodels} is the process of optimizing the output of a large language model and it references an authoritative knowledge base outside of its training data sources before generating a response. This paradigm has gained traction for tasks requiring factual accuracy and up-to-date information, such as question answering, summarization, and dialogue generation.

\begin{wrapfigure}{r}{0.5\textwidth}
    \vspace{-10pt}  % (Optional) fine-tune vertical spacing
    \centering
    \includegraphics[width=0.48\textwidth]{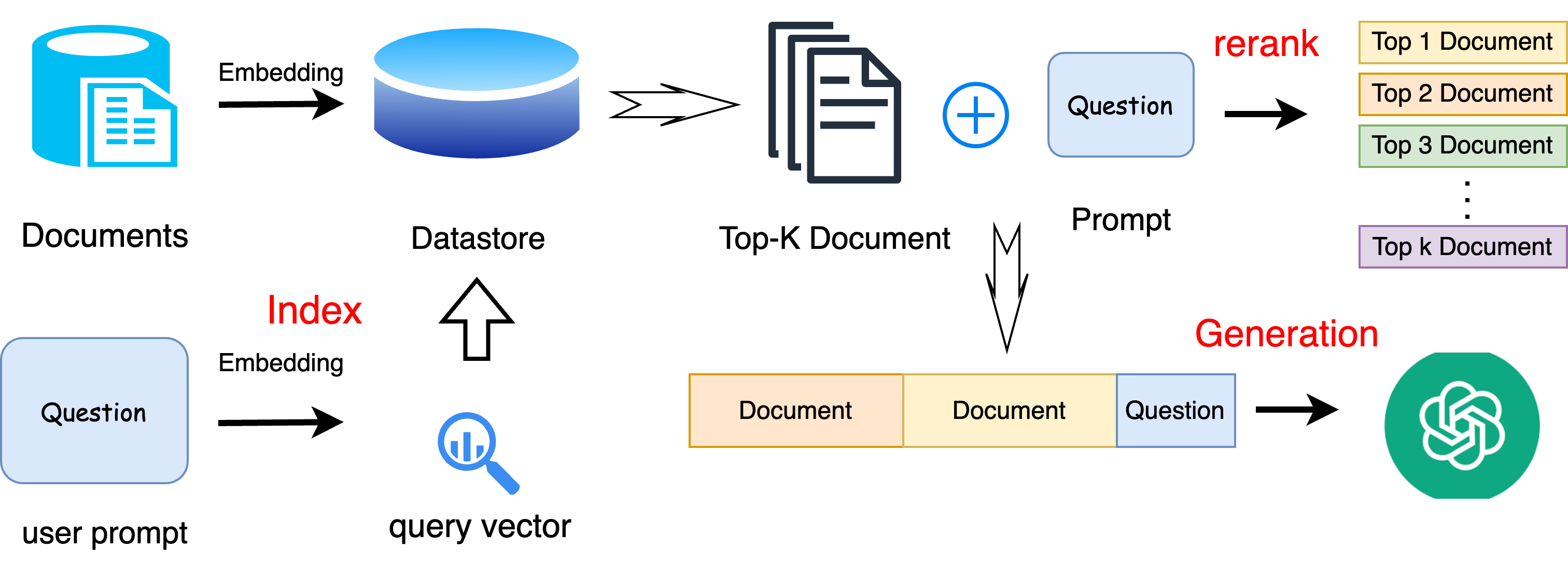} % or your figure file
    \caption{Classic RAG Workflow:
        The query is embedded and used to retrieve top-K documents. Then the reranker selects the most relevant ones which are combined with the query to generate the final response.
    }
    \label{fig:myfig}
    \vspace{-10pt}  % (Optional) fine-tune vertical spacing
\end{wrapfigure}

\begin{figure*}[t]
    \centering
\subfigure[TriviaQA]{\includegraphics[width=0.32\textwidth]{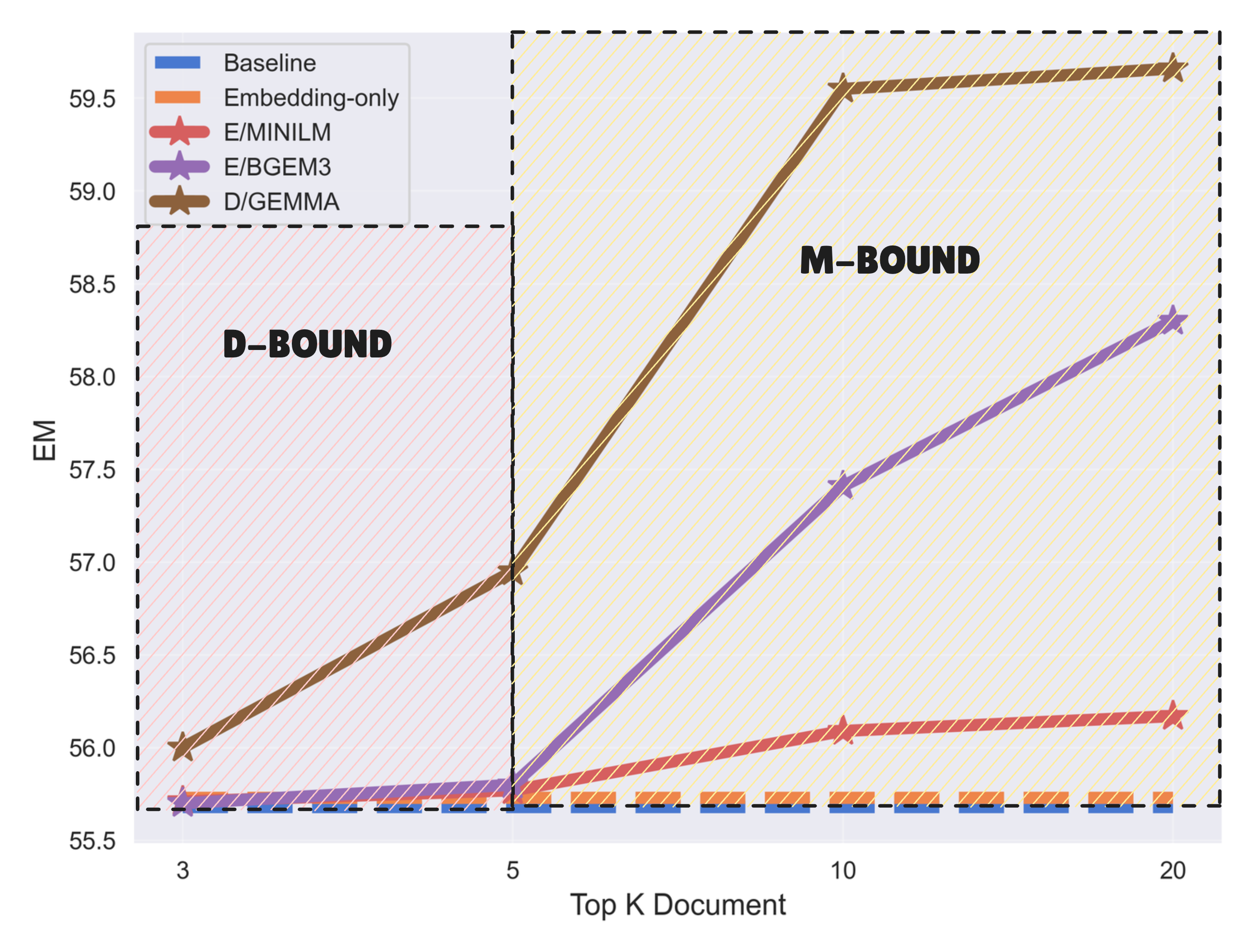}}
\subfigure[NaturalQA]{\includegraphics[width=0.32\textwidth]{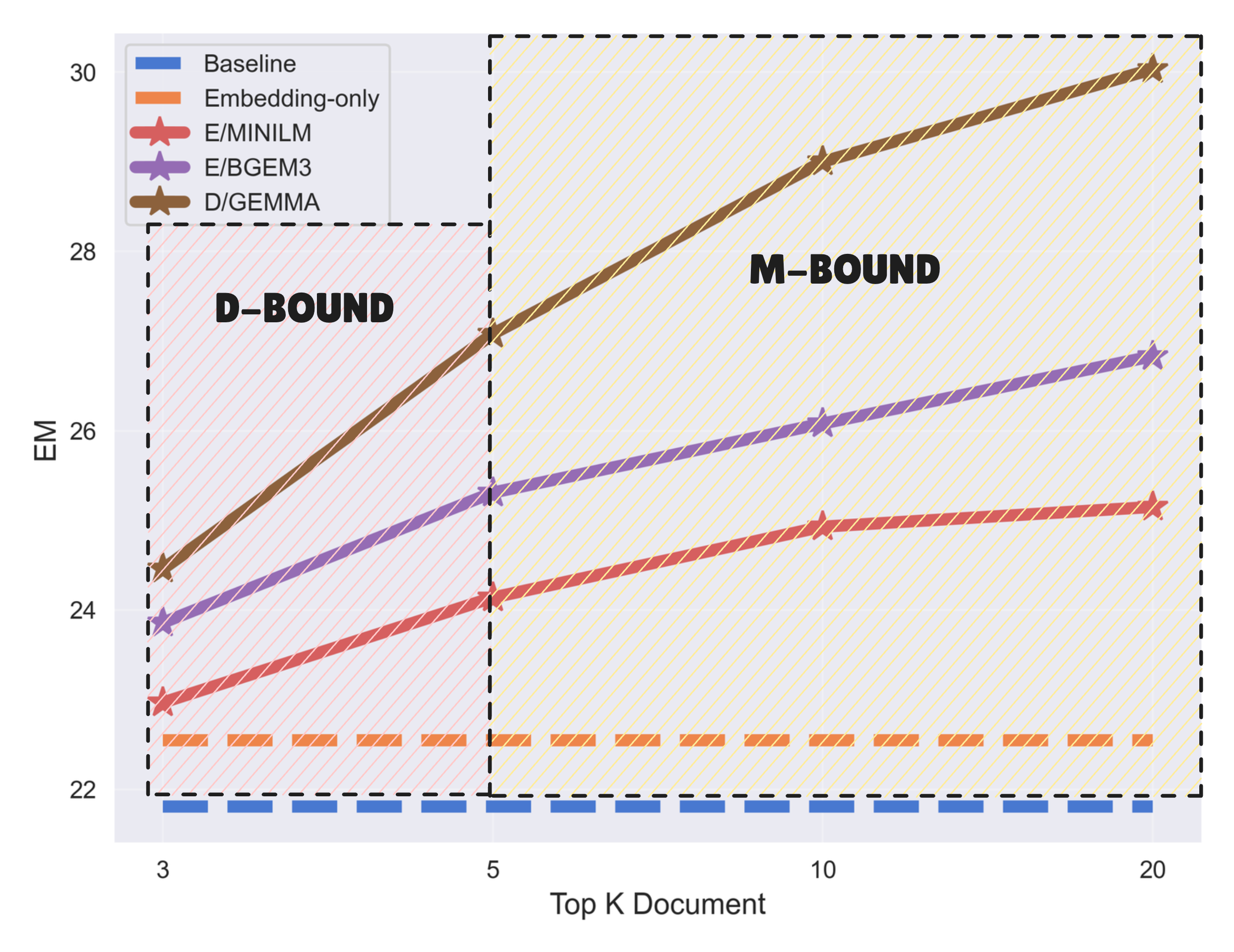}}
\subfigure[PopQA]{\includegraphics[width=0.32\textwidth]{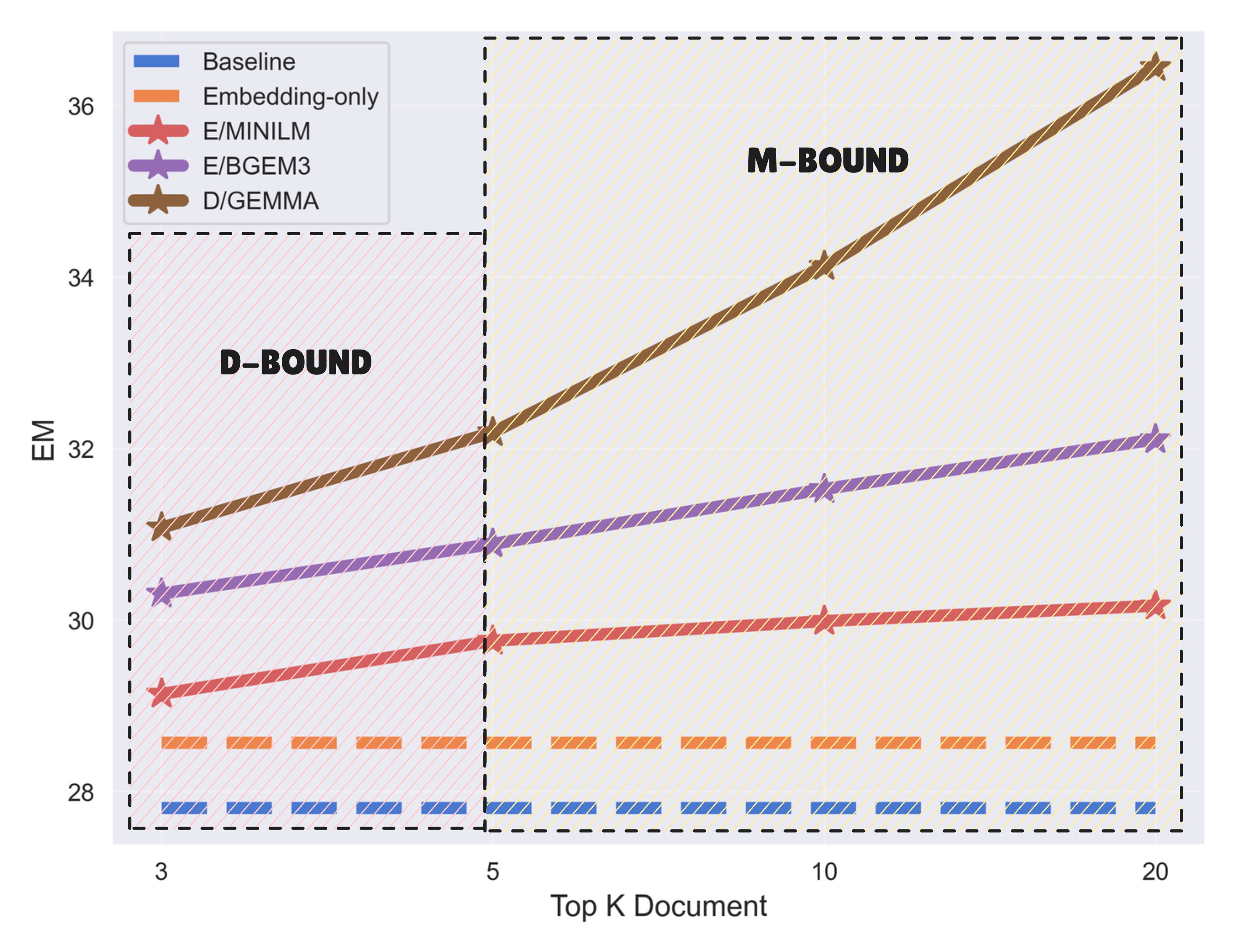}}
    \caption{RAG downstream performance with different rerankers: Subfigures a, b, and c show the performance curves of exact match (EM) scores on TriviaQA \citep{joshi2017triviaqalargescaledistantly}, NaturalQA \citep{nq}, and PopQA \citep{popqa} with various rerankers. The x-axis denotes the number of retrieved documents involved during reranking from which the top-1 document is selected for generation. \textbf{D-BOUND} represents the performance upper bound is based on the number of documents during rerank while \textbf{M-BOUND} reflects that the upper bound is determined by the reranker’s ability to identify the most relevant document. The generation model is \texttt{meta-llama/Llama-3.1-8B-Instruct}. The five labels represent different configurations:
Baseline (No RAG),
Embedding-only (retrieves the top document directly using cosine similarity),
E/MINILM (uses the ms-marco-MiniLM-L6-v2 reranker which is Encoder-only \citep{cross-encoder-ms-marco-MiniLM-L6-v2}),
E/BGEM (uses the bge-reranker-v2-m3 reranker which is Encoder-only \citep{bge-reranker-v2-m3}), and
D/GEMMA (uses the Gemma 2B reranker which is Decoder-only \citep{bge-reranker-v2-gemma}).}
    \label{fig:empower}

\end{figure*}

Traditional RAG pipelines rely on retrievers which depend on embedding representations and cosine similarity to fetch relevant documents \citep{bm25, reimers-2019-sentence-bert, wang2024textembeddingsweaklysupervisedcontrastive}. While this approach is straightforward, it often struggles to achieve optimal results in more complex scenarios. To solve the problem, advancements have introduced reranker mechanisms that refine the retrieved documents to improve relevance and contextuality before generation. These rerankers, often transformer-based, significantly boost the quality of generated content by ensuring the most pertinent documents are prioritized.

Early Rerankers were predominantly trained on encoder-only models such as BERT \citep{devlin2019bertpretrainingdeepbidirectional} or XLM-RoBERTa \citep{conneau2020unsupervisedcrosslingualrepresentationlearning}, leveraging their strong encoding capabilities to improve retrieval precision. However, recent advancements have demonstrated the growing dominance of decoder-based rerankers which capitalize on the powerful generative language capabilities of modern decoder models \citep{chen2024bgem3embeddingmultilingualmultifunctionality, ma2023finetuningllamamultistagetext, pradeep2023rankzephyr}. By fine-tuning these decoder-based models on tasks originally designed for encoder-only rerankers, they achieve significant gains in performance, benefiting from both their inherent generative power and the fine-grained contextual understanding acquired during fine-tuning. 

% However, the added computational steps increase time-to-first-token (TTFT), posing challenges for latency-sensitive applications.

\subsection{KV-cache Reuse of LLM}
In generation scenario,  KV-cache reuse has emerged as a vital optimization technique for improving the efficiency of large language models (LLMs) during inference. \citep{liu2024cachegenkvcachecompression}. % particularly in RAG scenarios~\cite{??}.
In RAG scenarios, prior works mostly focus on reusing the KV-cache of the retrieved documents to speed up the generation stage \citep{yao2024cacheblend, gim2024prompt, jin2024ragcache}. %for instance, reusing the KV cache of the retrieved documents to speed up the generation stage~\cite{promptcache, cacheblend}, 
In this work, we focus on reusing KV-cache during the retrieval stage to speed up the rerank process.
%By storing and reusing precomputed KV caches during the pre-filling stage, this approach significantly reduces redundant computations, leading to substantial improvements in time-to-first-token (TTFT) and serving throughput~\cite{??}. 
%This optimization minimizes overhead while maximizing the efficiency of inference workflows. \yw{add ref and make it longer}
\section{Observation}

In this section, we highlight the key observations that inspired the design of HyperRAG. In Section \ref{sec:trade-off}, we present the quality-efficiency trade-off curve inherent in RAG serving systems which our design aims to address. This trade-off is primarily influenced by the computational cost of embedding retrieval and reranking. In Section \ref{sec:reuse}, we explore the benefits of KV cache reuse during reranking inference and explain why reranking is the optimal scenario for leveraging KV cache reuse.

\subsection{Reranker Empowers RAG}

In retrieval-augmented generation (RAG), the quality of the retrieved documents directly influences the final generation output. However, the initial retrieval step often returns a set of candidate documents with mixed relevance. A reranker plays a critical role in reordering these candidates to identify the most relevant ones, enabling the model to generate more accurate and grounded responses.

Our findings show that a powerful reranker is essential to fully realize the potential of RAG. Without it, even high-capacity language models struggle to make effective use of the retrieved context. As shown in Figure~\ref{fig:empower}, increasing the number of candidates reranked improves the performance ceiling (D-BOUND), but only a strong reranker can consistently select the best document and push the system closer to the model's generation upper bound (M-BOUND). This observation highlights the reranker's pivotal role in bridging retrieval and generation and demonstrates that a RAG system can fully realize its potential only when equipped with a powerful reranker.

\begin{figure}[t]
    \centering
    \subfigure[Latency vs EM]{
        \includegraphics[width=0.22\textwidth]{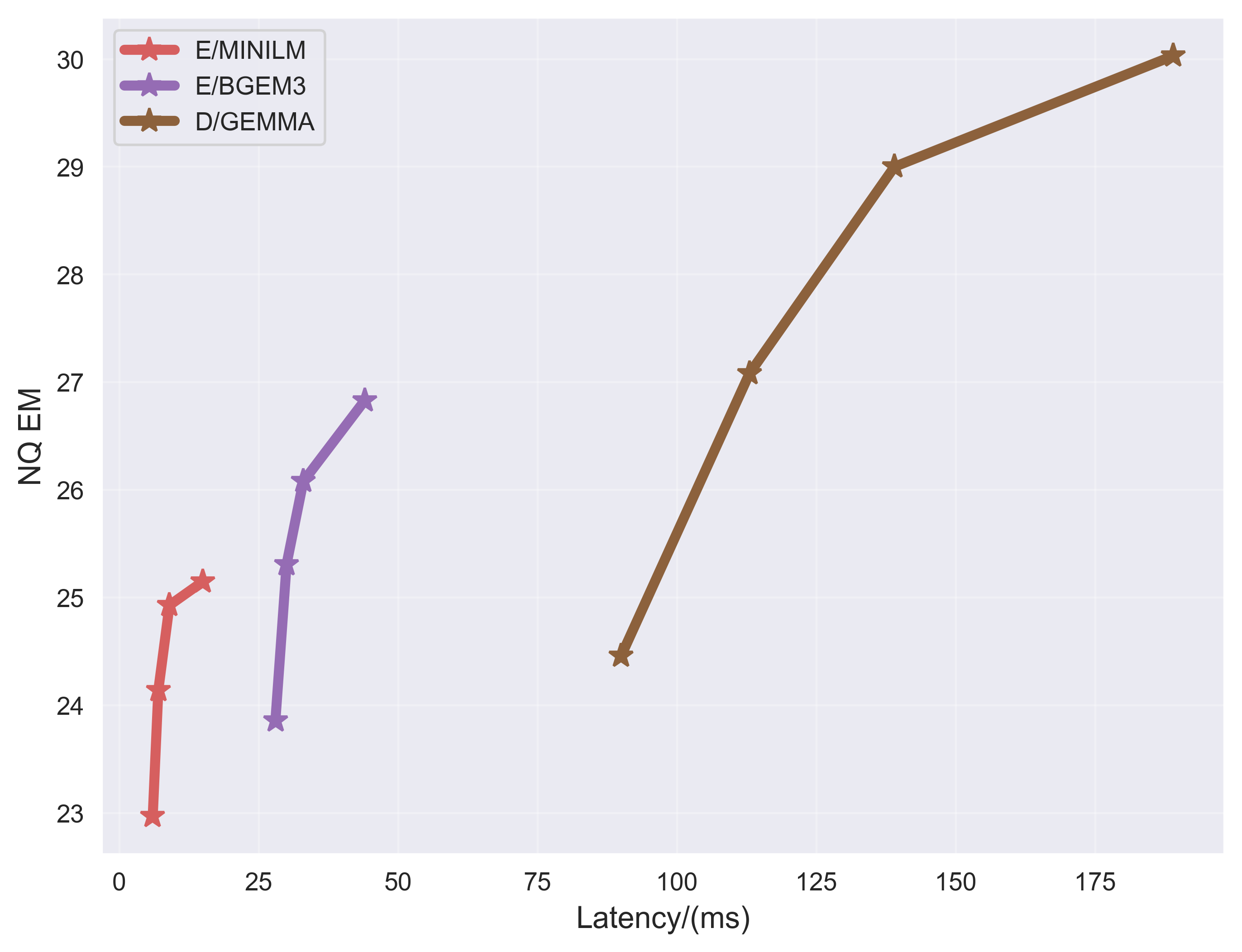}
        \label{fig:reuse_a}
    } 
    \subfigure[Full vs Reuse]{
        \includegraphics[width=0.23\textwidth]{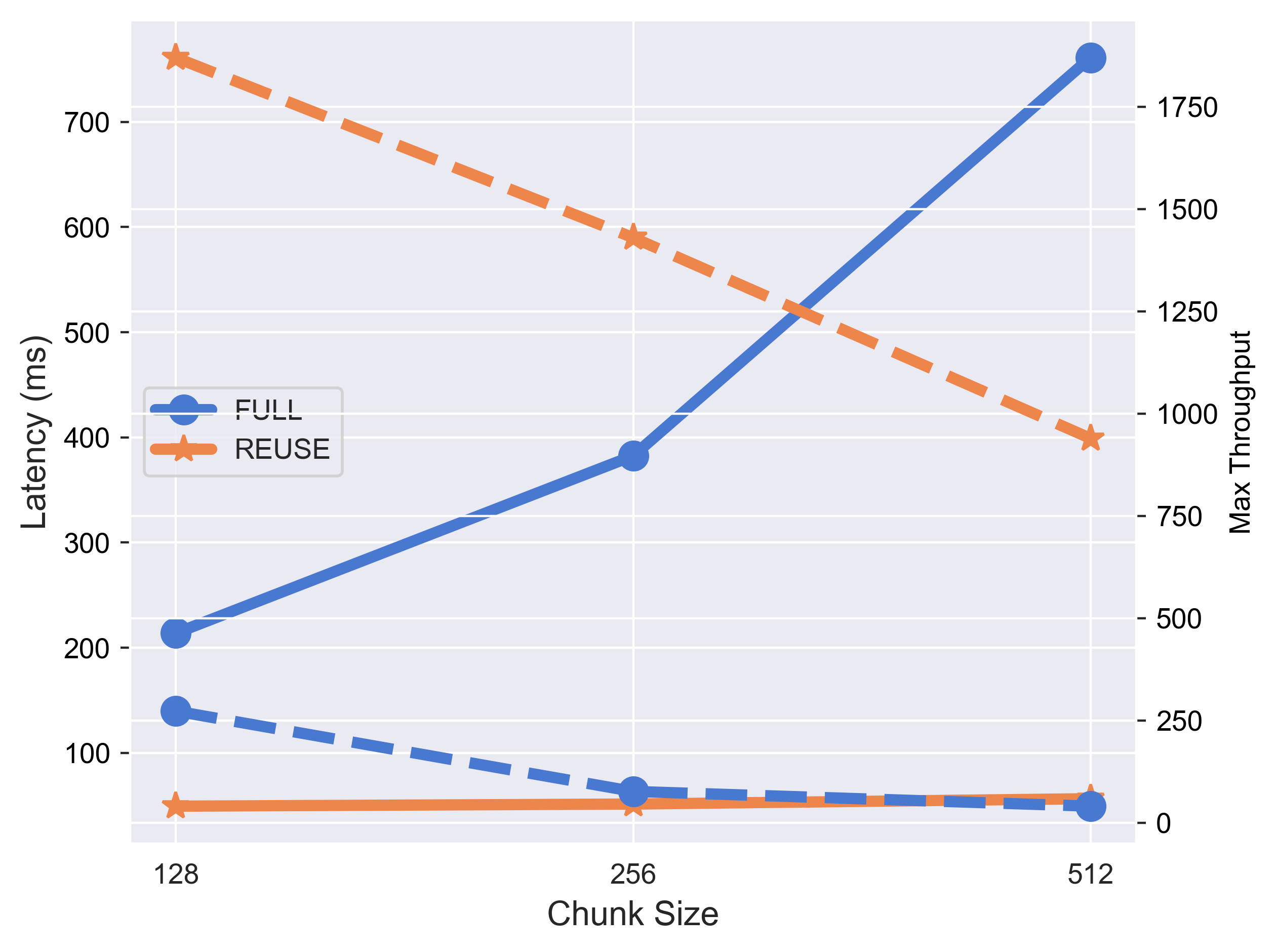}
        \label{fig:reuse_b}
    } 
    \subfigure[Memory Footprint]{
        \includegraphics[width=0.22\textwidth]{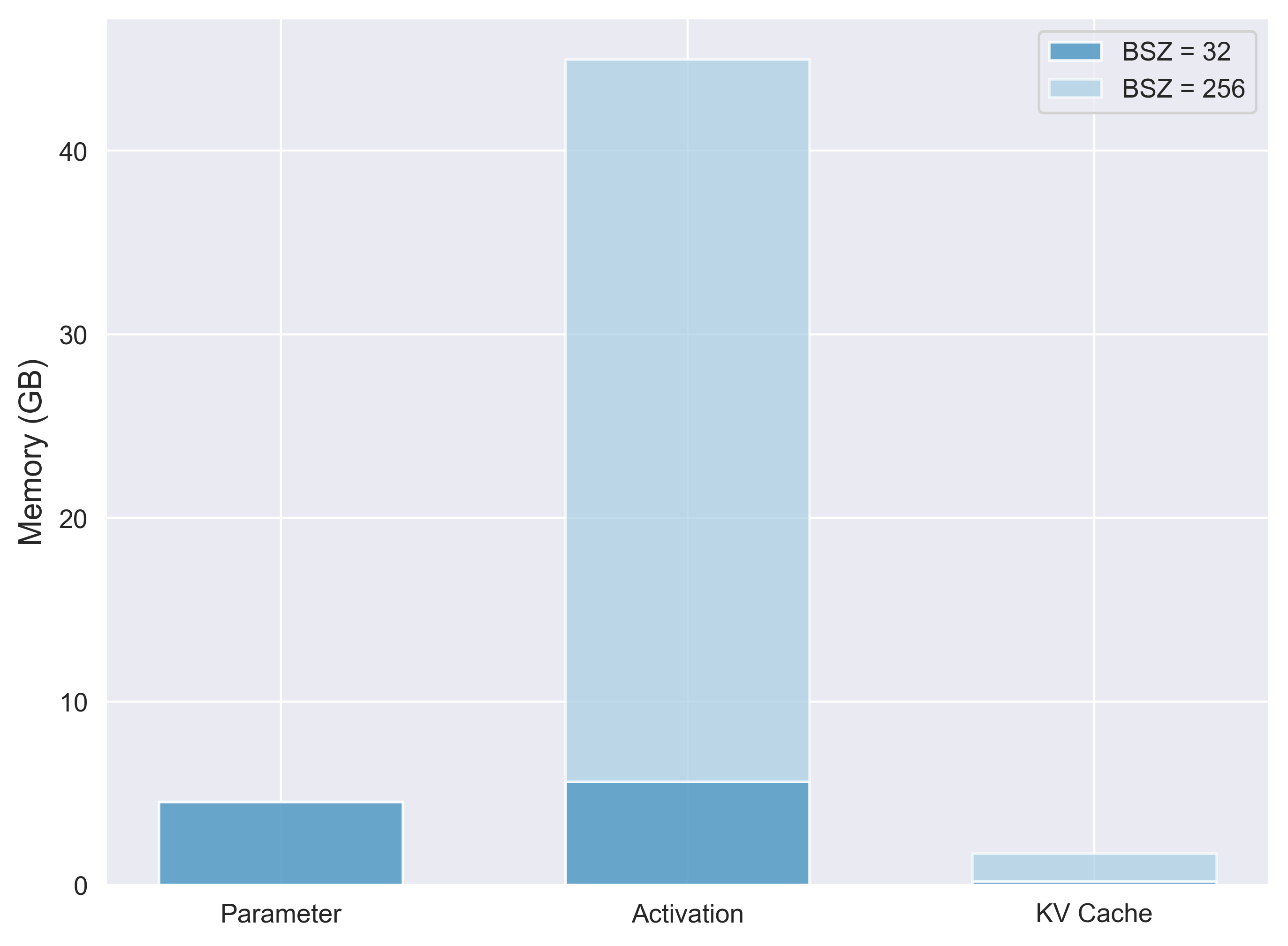}
        \label{fig:reuse_c}
    }    
    \subfigure[Throughput vs batch]{
        \includegraphics[width=0.22\textwidth]{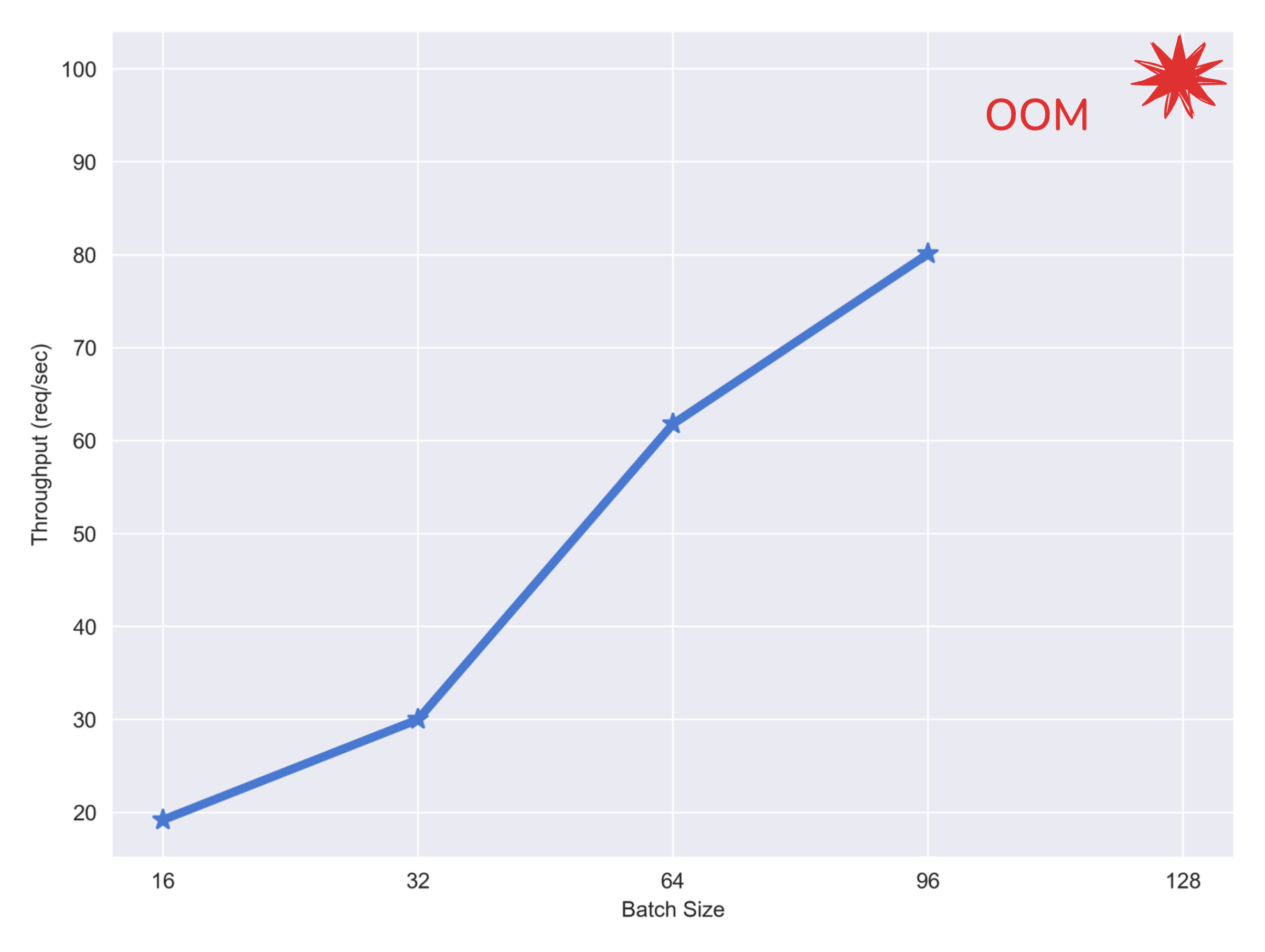}
        \label{fig:reuse_d}
    }
    
\caption{Efficiency Observations for the Reranker:  
Subfigure~\ref{fig:reuse_a} illustrates the trade-off between latency and NaturalQA performance across different reranker models.  
Subfigure~\ref{fig:reuse_b} presents the latency and throughput of the Gemma-2B reranker under varying document chunk sizes, with the query chunk size fixed at 48. The blue line indicates full computation, while the orange line represents computation with KV-cache reuse. Solid lines denote latency, and dashed lines denote throughput.  
Subfigure~\ref{fig:reuse_c} shows the memory footprint of the Gemma-2B reranker during inference with different batch sizes, using a fixed input length of $256 + 48 = 304$.  
Subfigure~\ref{fig:reuse_d} highlights how throughput increases with larger batch sizes up until an out-of-memory (OOM) error occurs for Gemma-2B reranker model inference.}

    \label{fig:reuse}
\end{figure}

\subsection{Quality-Efficiency Trade-off}
\label{sec:trade-off}
While larger rerankers deliver better generation quality, they are typically built on large language models and incur significantly higher latency. As shown in Figure~\ref{fig:reuse_a}, more powerful rerankers such as Gemma-2B reranker model achieve higher EM scores on NaturalQA but at a much greater computational cost compared to lighter-weight encoder models.

This trade-off becomes particularly problematic in real-world deployment scenarios. For each incoming query, the reranker must evaluate all top-$k$ retrieved documents by pairing the query with each candidate and processing them through the model. To improve throughput, multiple queries are typically batched together within the RAG service pipeline, resulting in exceptionally large batch sizes. Under these conditions, rerankers based on large language models suffer from severe performance degradation compared to their smaller counterparts. This efficiency bottleneck is a key reason why many production RAG systems avoid using these more powerful rerankers despite their superior retrieval quality.

% \begin{figure}[]
%     \centering
%     \includegraphics[width=1.0\linewidth]{section/image/mask.drawio.png}
%     \caption{KV Cache Representation in the Attention Layer: The grid represents the key-value pairs, where rows correspond to sequence positions of documents and queries, and columns to token embeddings. The color coding highlights the separation: white cells represent masked positions, red cells represent the document embeddings, and blue cells represent the query embeddings. The process involves loading document KV pairs from SSD storage into memory (red cells) and computing query KV pairs on the GPU (blue cells) for attention calculations. }
%     \label{fig:reuse mask}
% \end{figure}

\subsection{KV-Cache Reuse}
\label{sec:reuse}

% \begin{table*}[]
% \centering
% \caption{KV Cache Memory Usage and Performance of rerankers (BF16 precision)}
% \label{tab:mla_models}
% \begin{tabular}{|l|c|c|c|}
% \hline
% \textbf{Model} & \textbf{Model Size} & \textbf{KV Memory/token} & \textbf{MS Marco} \\ \hline
% rankllama & 7B & 512 KB & 47.6 \\ \hline
% bge-reranker-large & 335M & 192 KB & 43.4 \\ \hline
% bge-reranker-gemini-2 & 2B & 144 KB & 45.2 \\ \hline
% mini-cpm-reranker & 4B & 35 KB & 44.6 \\ \hline
% \end{tabular}
% \end{table*}

Based on the above observation, the key challenge is to leverage powerful reranker models while mitigating the service efficiency degradation they introduce. Given the substantial computational overhead of large language model inference, reusing the KV-cache has emerged as a promising strategy for accelerating decoder-based models. While prior work has primarily focused on applying KV-cache reuse during the generation phase, we argue that extending this technique to the reranking stage is both more efficient and better aligned with its characteristics.

In this section, we outline several important \textbf{properties} of reranking that make it suitable for KV-cache reuse, and highlight the \textbf{potential} benefits this approach can bring in terms of efficiency improvements.

\textbf{Property 1: Lossless} Compared to encoder-based rerankers, decoder-based rerankers benefit from the tri-mask mechanism, which enables lossless two-stage inference. Specifically, this mechanism ensures that computing the score for a "document + query" pair yields the same result as first performing inference on the document alone and then processing the query using the precomputed KV-cache from the document.

Another key advantage is that reranker inference operates on a document-query pair basis, evaluating one document chunk against a query at a time. This pairwise structure eliminates a major challenge present in KV-cache reuse during the generation phase of RAG where different orderings of documents (e.g., \textit{Document A + Document B + Query} vs. \textit{Document B + Document A + Query}) result in different KV-cache and necessitate costly recomputation. In contrast, reranking avoids this issue entirely since each document is scored independently with respect to the query. 

\textbf{Property 2: Static} The chunk size of the document is often fixed, which brings significant benefits for KV-cache management and inference compilation. By maintaining a consistent chunk size, the KV-cache structure becomes predictable which simplifies memory allocation and reduce unnecessary padding.

\textbf{Property 3: Large Reuse Ratio} In the reranking scenario, the token length of document chunks is typically larger than that of the query. This results in a high reuse ratio, which significantly reduces latency since the majority of tokens can be reused without recalculation.

These favorable properties make reranking an especially suitable scenario for applying KV-cache reuse. Our observations further reveal the untapped potential of this approach to significantly improve efficiency without compromising accuracy.

\textbf{Potential 1: Less Computation} Thanks to the high reuse ratio enabled by KV-cache reuse, reranking with decoder-based models can significantly reduce redundant computation. Instead of recomputing document representations for each query, the cached document-side KV states can be reused across different document–query pairs.  Figure~\ref{fig:reuse_b} demonstrates this effect using the Gemma-2B reranker. As chunk size increases, the latency of full inference grows rapidly due to the increasing computation required for processing longer sequences. In contrast, with KV-cache reuse, latency remains nearly constant and substantially lower across all chunk sizes. This indicates that the document-side computation—normally the dominant contributor to latency—has been effectively amortized. Naturally, the throughput of reranking also improves as a result of the reduced latency.

\textbf{Potential 2: Lower Memory Footprint Enables Larger Batches}
KV-cache reuse also offers substantial memory savings, which can translate into significant throughput improvements. Figure~\ref{fig:reuse_c} shows the memory footprint of the prefill stage when running the Gemma-2B reranker. As the chunk size increases, the intermediate activations during full prefilling become large and quickly exhaust available GPU memory. This limits the maximum batch size that can be processed concurrently. Figure~\ref{fig:reuse_d} demonstrates that batch size can be a key bottleneck for throughput—when memory is constrained, which means that the memory upper bound comes first compared to compuation upper bound.

In contrast, KV-cache reuse significantly reduces the intermediate memory requirements by skipping document-side prefilling during inference. With reduced memory usage, the reranker can support higher levels of parallelism, enabling larger batch sizes and increased throughput during reranking.

\section{HyperRAG}
\begin{figure*}[t]
    \centering
    \includegraphics[width=0.93\linewidth]{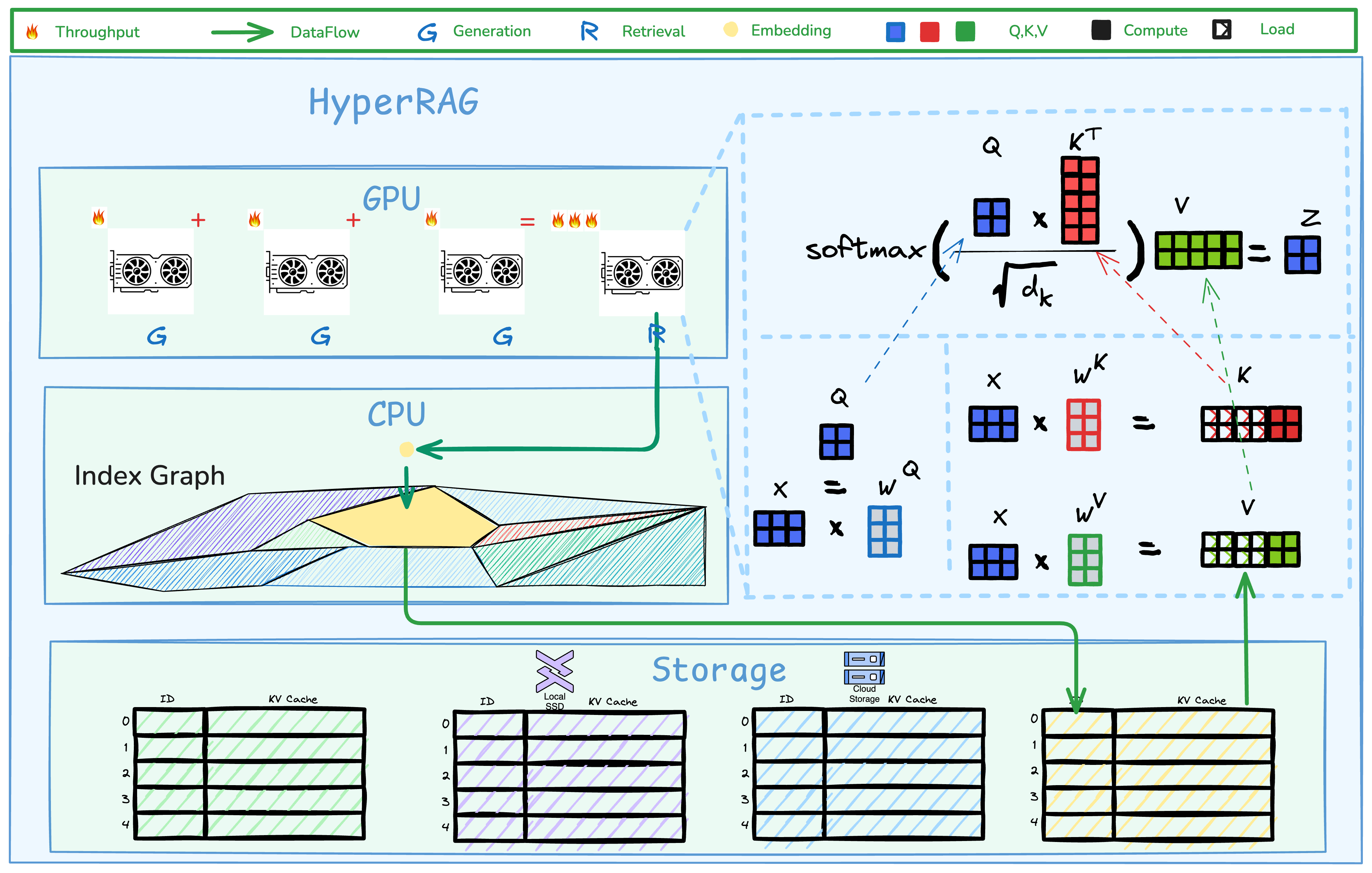}
    \caption{Overview of HyperRAG}
    \label{fig:enter-label}
\end{figure*}

In this section, we present the overall system design of \textbf{HyperRAG}, our high-performance retrieval-augmented generation (RAG) framework optimized for both efficiency and scalability. 
The rest of this section is organized as follows. Section~\ref{sec:kv-op} introduces the design and benefits of KV-cache compression in HyperRAG. Sections~\ref{sec:gpu}, \ref{sec:cpu}, and \ref{sec:storage} then discuss the hierarchical design of HyperRAG, moving from GPU to CPU and finally to Storage.

% In section \ref{}

\subsection{KV-cache Compression: Addressing the Storage Bottleneck}
\label{sec:kv-op}
While KV-cache reuse offers significant latency and throughput benefits, it introduces a new set of challenges, particularly when scaling to large datastores. In this scenario, the system may need to offload KV-cache to external storage devices such as NVMe or cloud-based object stores. As a result, the bottleneck shifts from GPU computation to the storage and data transfer layers. Two major issues emerge: the overall size of the KV-cache and the bandwidth required to transfer it between storage and GPU memory. To address this, we propose lightweight KV-cache compression techniques that are critical for scaling HyperRAG to large deployments.

\textbf{Efficient Attention Mechanisms:}  
Compared to traditional attention mechanisms, recent designs such as Grouped Query Attention (GQA) \citep{ainslie2023gqatraininggeneralizedmultiquery} and Multi-Latent Attention (MLA) \citep{deepseekai2024deepseekv2strongeconomicalefficient} significantly reduce KV-cache memory requirements per token. While originally developed to improve GPU decoding throughput, these mechanisms are particularly beneficial in the context of HyperRAG. Since reranker inference involves transferring KV-cache for every document–query pair, reducing the per-token KV footprint directly eases pressure on storage bandwidth and memory usage. Among the rerankers we evaluated, the Gemma-2B reranker is especially well-suited for this setting due to its low KV memory cost per token, making it ideal for large-scale and cache-intensive RAG inference.

\textbf{KV-cache Quantization:}  
Quantization \citep{han2016deepcompressioncompressingdeep, lin2024awqactivationawareweightquantization, xiao2024smoothquantaccurateefficientposttraining} is a widely-used technique for compressing model weights and activations by reducing precision. In HyperRAG, we extend this concept to KV-cache, focusing on minimizing storage and transfer cost while preserving model performance. Rather than applying full-model quantization schemes like \textbf{W8A8} or \textbf{W4A16}, we adopt specialized formats such as \textbf{KV8W16A16} and \textbf{KV4W16A16}, where quantization is applied exclusively to KV-cache while keeping model weights and activations in higher precision. However in the following experiment we do not use quantization further. Appendix \ref{app:quant} shows the reason why we do not use it for Gemma-2B reranker.

\subsection{GPU-Centric Computation}
\label{sec:gpu}
In HyperRAG, GPU resources are responsible for two main computational roles: (1) retrieval-side processing—including embedding generation and reranking—and (2) generation-side inference, where the top-ranked document is combined with the query to produce the final response.

\begin{wrapfigure}{r}{0.5\textwidth}
    \vspace{-10pt}  % (Optional) fine-tune vertical spacing
    \centering
    \includegraphics[width=0.48\textwidth]{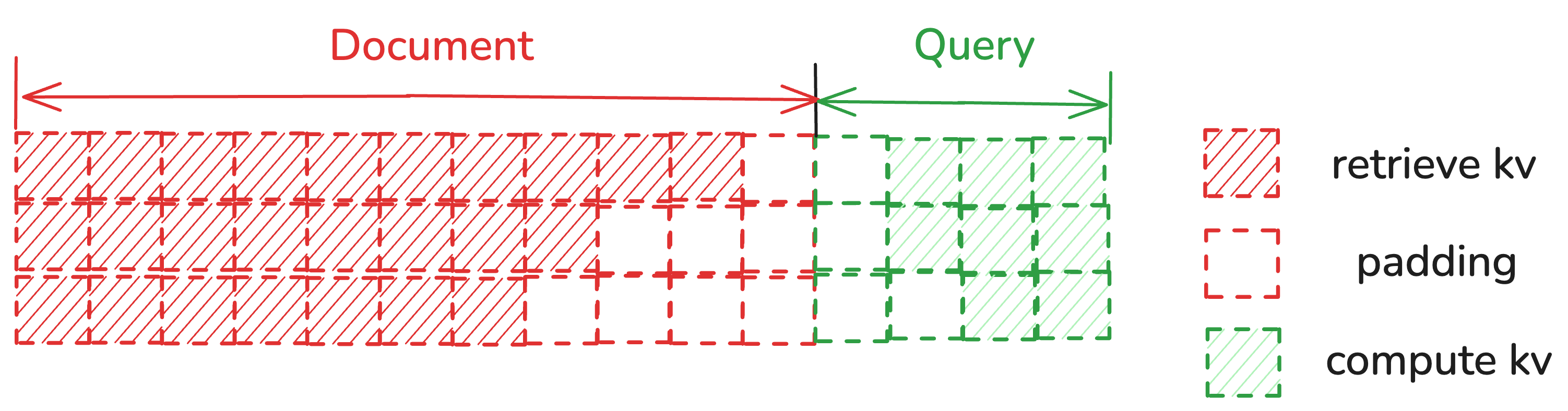} % or your figure file
    \caption{Static KV Layout: During reranking, we allocate a fixed-length KV buffer for attention. The buffer consists of a static document segment (retrieved KV, shown in red) and a static query segment (computed KV, shown in green). }
    \label{fig:static_attention}
    \vspace{-10pt}  % (Optional) fine-tune vertical spacing
\end{wrapfigure}

To maximize system throughput, it is crucial to balance the computational load between these two stages. If either the reranking or generation side becomes a bottleneck, overall throughput suffers due to underutilized GPU resources. HyperRAG addresses this by disaggregating retrieval and generation pipelines and carefully managing GPU allocation. By dynamically tuning the number of GPUs assigned to each stage and leveraging KV-cache reuse to reduce redundant computation, HyperRAG maintains a balanced and efficient end-to-end inference pipeline.

In addition to load balancing, HyperRAG further optimizes GPU execution by adopting a \textbf{static attention layout}, enabling better fusion and graph capture through torch compile and CUDA graphs \citep{torch}. As illustrated in Figure~\ref{fig:static_attention}, we fix the structure of the attention kv-input such that the document occupies a static prefix region while the query occupies a fixed-length suffix. The document tokens use precomputed KV-cache (retrieved from memory) and only the query portion contributes to new KV computation. This static partitioning allows the entire decoding process to be compiled ahead of time and executed with minimal runtime overhead, greatly improving inference efficiency.

\subsection{CPU-Centric Index}
\label{sec:cpu}
In the Embedding retrieval stage of HyperRAG, we leverage FAISS, a highly optimized library for efficient similarity search on dense vectors. This CPU-centric approach ensures the rapid indexing and retrieval of top-K documents. FAISS operates on the dense embeddings generated by the model, implementing approximate nearest neighbor (ANN) search algorithms that are both scalable and precise. Specifically, we employ IVF (Inverted File Index) and HNSW (Hierarchical Navigable Small World) \citep{malkov2018efficientrobustapproximatenearest} indexing structures within FAISS \citep{douze2025faisslibrary} which enables fast and memory-efficient search while maintaining high recall. 

\subsection{Storage Backend}
\label{sec:storage}
To support efficient and scalable KV-cache access in HyperRAG, we build our storage backend on top of the open-source \textit{LMCache} framework \citep{cheng2024large}. This backend is designed to provide near-constant KV-cache retrieval latency, even as the total cache size scales to billions of tokens. It offers fast and reliable KV-cache management across multiple storage backends, such as local NVMe drives and remote systems like Redis.

A key feature of our storage layer is its ability to seamlessly partition the KV-cache across multiple storage devices, such as independent NVMe drives or remote object storage. Given the scale of KV-cache required for large datastores, such partitioning is essential. To maintain retrieval efficiency in this setting, we align the partitioning strategy with the indexing structure of the embedding retriever. Specifically, we group KV-cache entries by the \textit{coarse centroid ID} used in the FAISS IVF index. This means that all KV entries assigned to the same coarse cluster are stored within the same storage backend.

This design offers a critical performance benefit: since approximate nearest neighbor (ANN) search during retrieval only probes a small number of centroid clusters, the corresponding KV-cache retrieval is limited to a single backend. As a result, we avoid cross-device fetching, reduce latency, and improve I/O locality.

\section{Experiment}
In this section, we present the results of the benchmark and performance experiments.

\begin{table*}[t]
\centering
\begin{tabular}{c|c|ccc|ccc|ccc}
\toprule
\multirow{2}{*}{\centering C} &  \multirow{2}{*}{\centering Metrics} & \multicolumn{3}{c|}{\textbf{BGE-M3}} & \multicolumn{3}{c|}{\textbf{Gemma}} & \multicolumn{3}{c|}{\textbf{H-Gemma}}\\
   \cmidrule(lr){3-5} \cmidrule(lr){6-8} \cmidrule(lr){9-11}
 &  & \textbf{5} & \textbf{10} & \textbf{20} & \textbf{5} & \textbf{10} & \textbf{20} & \textbf{5} & \textbf{10} & \textbf{20} \\
\midrule
\multirow{4}{*}{P} & TriviaQA & 55.80 &	57.41&	58.30& 56.95&	59.55&	59.66& 56.95&	59.55&	59.66\\
                        & NQ & 25.31 & 26.08 & 26.83 & 27.08 & 29.00 & 30.03 & 27.08 & 29.00 & 30.03 \\
                        & PopQA & 30.89 & 31.53& 32.11& 32.19 &34.13 & 36.45 & 32.19 &34.13 & 36.45 \\
                        & Throughput & 165.3 & 124.7 & 67.8 & 51.2 & 27.4 & 13.3 & 78.8 & 50.9 & 28.9 \\
\midrule
\multirow{4}{*}{D} & TriviaQA & 56.15 & 56.92 & 57.33 & 56.91 & 58.77 & 59.05 & 56.91 & 58.77 & 59.05 \\
                        & NQ & 25.13 & 25.87 & 26.60 & 26.55 & 28.47 & 29.50 & 26.55 & 28.47 & 29.50 \\
                        & PopQA & 29.98 & 30.62 & 31.21 & 29.98 & 32.02 & 34.21 & 29.98 & 32.02 & 34.21 \\
                        & Throughput & 75.8 & 69.0 & 41.7 & 28.4 & 15.2 & 7.5 & 48.7 & 36.2 & 21.1 \\
\bottomrule
\end{tabular}
\caption{Downstream Performance Evaluation:
We report Exact Match (EM) scores on three QA benchmarks along with throughput across different settings. The column labeled C denotes the corpus type: Passage-level (P) or Document-level (D). For each reranker model (BGE-M3 \citep{bge-reranker-v2-m3}, Gemma \citep{bge-reranker-v2-gemma}, HyperRAG Gemma(H-Gemma)), we vary the number of retrieved documents involved in reranking (shown as 5, 10, and 20).}
\label{tab:reranker_results}
\end{table*}

% \begin{table*}[t]
% \centering
% \begin{tabular}{c|c|ccc|ccc|ccc|}
% \toprule
% \multirow{2}{*}{\centering Corpus} &  \multirow{2}{*}{\centering Metrics} & \multicolumn{3}{c|}{\textbf{BGE-M3}} & \multicolumn{3}{c|}{\textbf{GEMMA}} & \multicolumn{3}{c|}{\textbf{H-GEMMA}}\\
%    \cmidrule(lr){3-5} \cmidrule(lr){6-8} \cmidrule(lr){9-11}
%  &  & \textbf{5} & \textbf{10} & \textbf{20} & \textbf{5} & \textbf{10} & \textbf{20} & \textbf{5} & \textbf{10} & \textbf{20} \\
% \midrule
% \multirow{4}{*}{Passage} & TriviaQA & - & - & - & - & - & - & - & - & - \\
%                         & NQ & - & - & - & - & - & - & - & - & - \\
%                         & PopQA & - & - & - & - & - & - & - & -& - \\
%                         & Throughtput & - & - & - & - & - & - & - & -& - \\
% \midrule
% \multirow{4}{*}{Document} & TriviaQA & - & - & - & - & - & - & - & - & - \\
%                         & NQ & - & - & - & - & - & - & - & - & - \\
%                         & PopQA & - & - & - & - & - & - & - & -& - \\
%                         & Throughtput & - & - & - & - & - & - & - & -& - \\
% \bottomrule
% \end{tabular}
% \caption{Performance comparison across reranker types, retrieval top-$k$, and dataset granularity.}
% \label{tab:reranker_results}
% \end{table*}

\subsection{Experiment settings}
\textbf{Corpus}:We evaluate HyperRAG under two different levels of retrieval granularity: passage-level and document-level.

\textit{Passage-Level Corpus:}  
For the passage-level setting, we use the psgs\_multiset-100 dataset as our base corpus. To ensure consistent input lengths and improve KV cache reuse efficiency, we rechunk all passages to a fixed size of approximately 200 words per chunk. In this case document tokens length is fixed as 256.

\textit{Document-Level Corpus:}  
For the document-level setting, we adopt the MS MARCO \citep{bajaj2018msmarcohumangenerated} document dataset. Each document is chunked into fixed-length segments with a maximum word length of 450. In this case document tokens length is fixed as 512.

\textbf{Metrics}:
To evaluate downstream performance, we utilized TriviaQA \citep{joshi2017triviaqalargescaledistantly}, NaturalQA \citep{nq}, and PopQA \citep{popqa} as benchmarks for performance evaluation. For the benchmark measurement, we collect the average throughput of RAG service(request per second) after warm-up for efficiency evaluation. 

\textbf{Model}:
For the embedding model of index search, we used the \texttt{contriever} \citep{izacard2022unsuperviseddenseinformationretrieval} model. In the generation stage, we employed the \texttt{Llama3.1-8B-Instruct} model \citep{llamaherd}.

% \textbf{Baseline}:
% We compared FastRaG with recent popular RaG workflow including LangChain, RaGFlow and FlashRAG. We also include the system with no RaG as baseline.

\textbf{Hardware}
The experiments were conducted on a system equipped with 8 NVIDIA A100 GPUs. For storage backend we use redis remote backend(4TB) and local nvme backend(4TB). We store all the hot KV-cache that will be used during inference on the QA dataset and then fulfill the whole storage with random sampling.

\subsection{Trade-off Performance}
We conduct experiments to evaluate both downstream performance and system efficiency. The results, presented in Table~\ref{tab:reranker_results}, demonstrate that HyperRAG maintains high-quality generation while achieving a 2–3× improvement in service throughput on average. These findings highlight HyperRAG's ability to effectively balance generation quality and cost-efficiency, making it a robust and practical solution for RAG service.

\section{Discussion}
In this section we will provide some key discussions around HyperRAG.
\subsection{RAG Service Framework}
One of the recent hot topics in RAG research is enabling the generation model to better identify and utilize the most relevant retrieved documents \citep{asai2023selfraglearningretrievegenerate, wei2025instructraginstructingretrievalaugmentedgeneration}.  HyperRAG aligns with this research direction, but with a key distinction: instead of relying solely on the generation model to perform this filtering, we delegate the relevance discrimination task to a powerful reranker. This reranker operates with similar mechanisms to large language models but is significantly more efficient due to its smaller size and the support of HyperRAG’s architecture. Moreover, this separation of relevance extraction offers greater flexibility during deployment.

\subsection{Real-World Application}

Storing reranker KV caches introduces a significant storage requirement. For example, we estimate that storing the full KV cache for the Gemma-2B reranker over the MS MARCO document dataset would require more than 40TB of storage. This raises questions about the practicality of deploying HyperRAG in real-world settings. However, we argue that HyperRAG remains a viable and efficient solution for two main reasons.

\begin{wrapfigure}{r}{0.5\textwidth}
    \vspace{-10pt}  % (Optional) fine-tune vertical spacing
    \centering
    \includegraphics[width=0.45\textwidth]{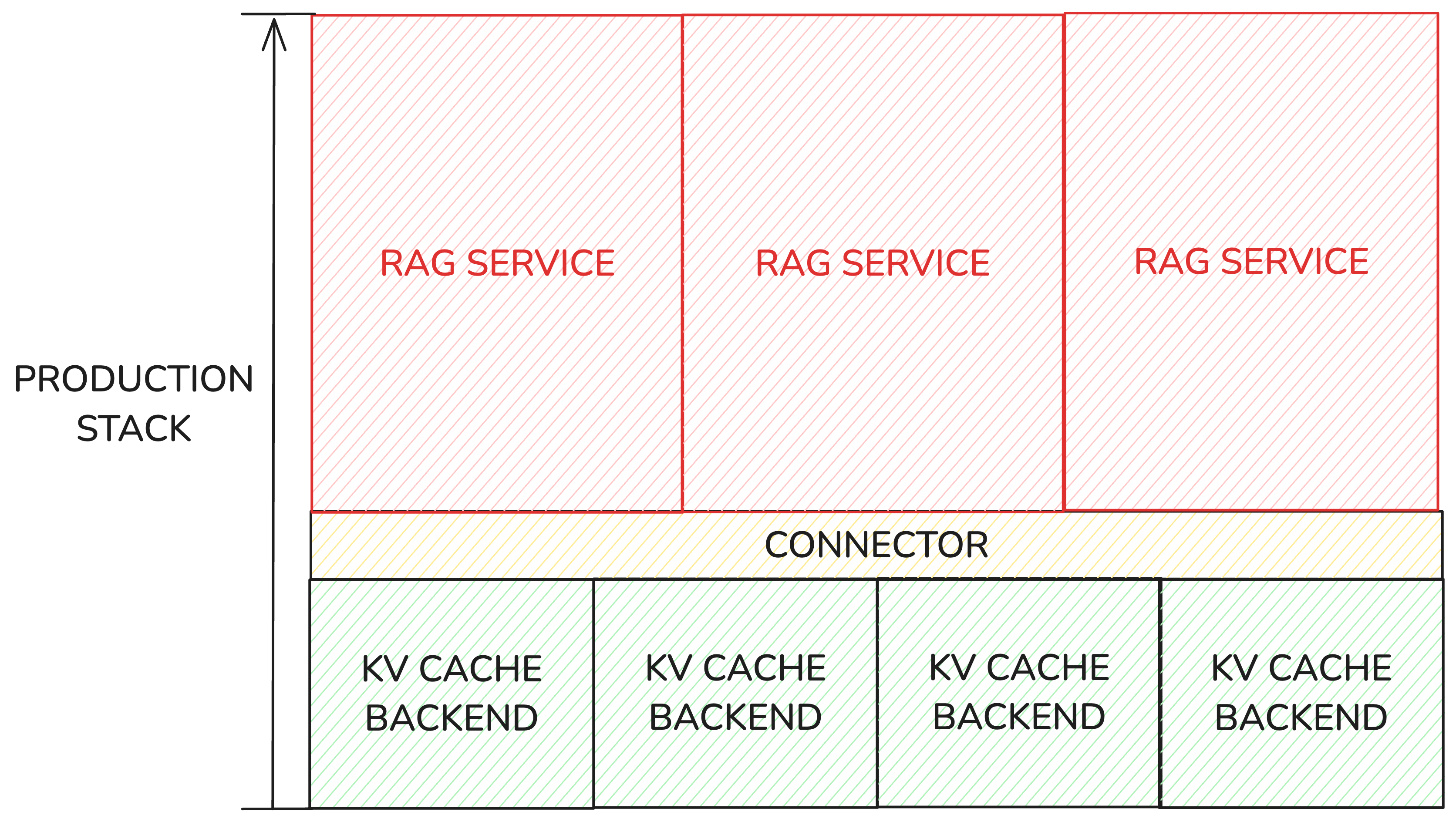} % or your figure file
    \caption{Production Stack: The stored KV cache backend can be shared across multiple RAG service engines. 
    }
    \label{fig:production}
    \vspace{-10pt}  % (Optional) fine-tune vertical spacing
\end{wrapfigure}

\textbf{Production Stack Design:}  
Figure~\ref{fig:production} illustrates a potential distributed deployment architecture for HyperRAG. In this setup, the KV cache backend functions as a shared memory store accessible by multiple RAG service engines. This design allows the storage burden to be amortized across multiple applications and services, enabling efficient and centralized knowledge sharing at scale.

\textbf{Financial Cost:}  
In practice, storage costs are negligible compared to GPU infrastructure. For instance, on Amazon Web Services, the cost of provisioning 8×A100 GPUs far exceeds that of acquiring 40TB of persistent storage. By introducing a relatively small additional cost for storage, HyperRAG achieves significantly higher throughput—making it a more cost-effective solution from the perspective of service providers aiming to optimize performance per dollar.

% \section*{Author Contributions}

% \section*{Acknowledgments}
% Use unnumbered first level headings for the acknowledgments. All

% \section*{Ethics Statement}
% Authors can add an optional ethics statement to the paper. 
% For papers that touch on ethical issues, this section will be evaluated as part of the review process. The ethics statement should come at the end of the paper. It does not count toward the page limit, but should not be more than 1 page. 

\bibliography{colm2025_conference}
\bibliographystyle{colm2025_conference}

\appendix
\section{Prompt Template}

Following \cite{jin2024longcontextllmsmeetrag}'s work, we utilize the prompt template for Generation Without RAG as follows:

\begin{verbatim}
<|begin_of_text|><|start_header_id|>system<|end_header_id|>

Answer the question based on your own knowledge. Only give me the answer
and do not output any other words.<|eot_id|><|start_header_id|>user<|end_header_id|>

Question: {question}<|eot_id|><|start_header_id|>assistant<|end_header_id|>
\end{verbatim}

The prompt template used for RAG is shown below:

\begin{verbatim}
<|begin_of_text|><|start_header_id|>system<|end_header_id|>

Answer the question based on the given document. Only give me the answer
and do not output any other words.
The following are given documents.

Doc {doc_id} (Title: {doc_title}) {doc_text}
Doc {doc_id} (Title: {doc_title}) {doc_text}
Doc {doc_id} (Title: {doc_title}) {doc_text}
Doc {doc_id} (Title: {doc_title}) {doc_text}
Doc {doc_id} (Title: {doc_title}) {doc_text}
Doc {doc_id} (Title: {doc_title}) {doc_text}

<|eot_id|><|start_header_id|>user<|end_header_id|>

Question: {question}<|eot_id|><|start_header_id|>assistant<|end_header_id|>
\end{verbatim}

\section{Quantization}
\label{app:quant}
To maintain compatibility with downstream operations (e.g., softmax), the KV cache is dequantized back to 16-bit precision after loading into GPU memory. This approach allows us to reduce I/O volume without introducing significant overhead during inference. However, in the HyperRAG we do not apply quantization because of two reasons. The first reason is around the downstream performance degradation. The second reason is that the original KV-cache of reranker model like Gemma-2B reranker is small enough. The introduction of quantization will introduce worse bandwidth performance.

\section{Reranker Finetune}

To improve reranking quality in our pipeline, we fine-tune all reranker models using a simple yet effective strategy: formatting the input as \texttt{[document] + [query]} rather than the original \texttt{[query] + [document]} format. This reversed input order aligns better with the decoder-based reranker architecture, which benefits from having the query appear later in the sequence—allowing it to attend over the full document context.

We use the open-source \texttt{FlagEmbedding} repository\footnote{\url{https://github.com/FlagOpen/FlagEmbedding}} as our fine-tuning framework. It supports a wide range of reranker backbones and provides efficient training tools. The rerankers are trained on the BGE-M3 training dataset, which contains multi-granularity positive and negative pairs curated for passage and document-level retrieval tasks. We follow the standard contrastive training setup, using positive and hard negative document pairs with a maximum sequence length of 512 tokens.

This fine-tuning strategy not only improves reranking performance but also makes the input format compatible with our static attention layout for KV cache reuse.
%%%%%%%%%%%%%%%%%%%%%%%%%%%%%%%%%%%%%%%%%%%%%%%%%%%%%%%%%%%%%%%%%%%%%%%%%%%%%%%
%%%%%%%%%%%%%%%%%%%%%%%%%%%%%%%%%%%%%%%%%%%%%%%%%%%%%%%%%%%%%%%%%%%%%%%%%%%%%%%

\end{document}